\documentclass[a4paper,orivec,runningheads]{llncs}
\usepackage{makeidx}  % allows for indexgeneration
\usepackage{graphicx}
\usepackage{subfigure}
\usepackage{amsmath}
\usepackage{amssymb}
\usepackage{verbatim}
\usepackage{algorithm}
\usepackage{algorithmic}
\usepackage{booktabs}
\usepackage{multirow}
\usepackage{pbox}
\usepackage{hyperref}
\usepackage{bm} 

\begin{document}

\pagenumbering{arabic}

%\frontmatter          % for the preliminaries
%

%
%\mainmatter              % start of the contributions
%
\title{Automatic Brain Tumor Segmentation using Convolutional Neural Networks with Test-Time Augmentation}

\titlerunning{Auto Brain Tumor Segmentation using CNNs with Test-Time Augmentation}  % abbreviated title (for running head)
%\author{aa,bb,cc}
%\author{Author Not Given}
%\author{Author Not Given}
%\author{No Author Given \\ ***  \\ ***}
\author{Guotai Wang\inst{1,2} \and Wenqi Li\inst{1,2} \and  S\'ebastien Ourselin\inst{1} \and Tom Vercauteren\inst{1,2}}

%\authorrunning{Guotai Wang, Maria A. Zuluaga, Rosalind Pratt, et al.} % abbreviated author list (for running head)
%%
%%%%% list of authors for the TOC (use if author list has to be modified)
%%\tocauthor{}
%%
%\institute{No Institute Given }
\institute{
$^1$School of Biomedical Engineering and Imaging Sciences, King's College London, London, UK \\
$^2$Wellcome / EPSRC Centre for Interventional and Surgical Sciences, University College London, London, UK \\
\email{guotai.wang@kcl.ac.uk}
%%\texttt{http://cmictig.cs.ucl.ac.uk}
}

\maketitle              % typeset the title of the contribution

\begin{abstract}
 Automatic brain tumor segmentation plays an important role for diagnosis, surgical planning and treatment assessment of brain tumors. Deep convolutional neural networks (CNNs) have been widely used for this task. Due to the relatively small data set for training, data augmentation at training time has been commonly used for better performance of CNNs. Recent works also demonstrated the usefulness of data augmentation at test time, in addition to training time, for achieving more robust predictions. We investigate how test-time augmentation can improve CNNs' performance for brain tumor segmentation. We used different underpinning network structures and augmented the image by 3D rotation, flipping, scaling and adding random noise at both training and test time. Experiments with BraTS 2018 training and validation set show that test-time augmentation can achieve higher segmentation accuracy and obtain uncertainty estimation of the segmentation results. 
\end{abstract}

\section{Introduction}
Gliomas are the most common primary brain tumors that start in the glial cells of the brain in adults. %They comprise about 30 percent of all brain and central nervous system tumors, and 80 percent of all malignant brain tumors~\cite{McKinsey2012}. 
They can be categorized according to their grade: Low-Grade Gliomas (LGG) exhibit benign tendencies and portend a better prognosis for the patient, while High-Grade Gliomas (HGG) are malignant and lead to a worse prognosis~\cite{Louis2016}. Medical imaging of brain tumors plays an important role for evaluating the progression of the disease before and after treament. Currently the most widely used imaging modality for brain tumors is Magnetic Resonance Imaging (MRI) with different sequences, such as T1-weighted, contrast enhanced T1-weighted (T1ce), T2-weighted and Fluid Attenuation Inversion Recovery (FLAIR) images. These sequences provide complementary information for different subregions of brain tumors~\cite{Menze2015}. For example, the tumor region and peritumoral edema can be highlighted in FLAIR and T2 images, and the tumor core region without peritumoral edema is more visible in T1 and T1ce images.  

Automatic segmentation of brain tumors and substructures from medical images has a potential for accurate and reproducible delineation of the tumors, which can help more efficient and  better diagnosis, surgical planning and treatment assessment of brain tumors~\cite{Menze2015,Bakas2017}. However, accurate automatic segmentation of the brain tumors is a challenging task for several reasons. First, the boundary between brain tumor and normal tissues is often ambiguous due to the smooth intensity gradients, partial volume effects, and bias field artifacts. Second, the brain tumors vary largely across patients in terms of size, shape, and localization. This prohibits the use of strong priors on shape and localization that are commonly used for robust segmentation of many other anatomical structures, such as the heart~\cite{Grosgeorge2013} and the liver~\cite{Wang2015c}. 

In recent years, deep Convolutional Neural Networks (CNNs) have achieved the state-of-the-art performance for multi-modal brain tumor segmentation~\cite{Wang17brats,Kamnitsas2017a}. As a type of machine learning approach, they require a set of annotated training images for learning. Compared with traditional machine learning approaches %such as support vector machines~\cite{Lee2005} and decision trees~\cite{Zikic2012}, 
they do not rely on hand-crafted features and can learn features automatically.  In~\cite{Havaei2016}, a CNN was proposed to exploit both local and global features for robust brain tumor segmentation. It replaces the final fully connected layer used in traditional CNNs with a convolutional implementation that obtains 40 fold speed up. This approach employs a two-phase training procedure and a cascade architecture to tackle difficulties related to the imbalance of tumor labels. Despite the better performance than traditional methods, this approach works on individual 2D slices without considering 3D contextual information.  DeepMedic~\cite{Kamnitsas2017} uses a dual pathway 3D CNN with 11 layers to make use of multi-scale features for brain tumor segmentation. For post-processing, it uses a 3D fully connected Conditional Random Field (CRF)~\cite{Krahenbuhl2011} that helps to remove false positives. DeepMedic achieved better performance than using 2D CNNs. However, it works on local image patches and therefore has a relatively low inference efficiency. In~\cite{Wang17brats}, a triple cascaded framework was proposed for brain tumor segmentation. The framework uses three networks to hierarchically segment whole tumor, tumor core and enhancing tumor core sequentially. It uses a network structure with anisotropic convolution to deal with 3D images, taking advantage of dilated convolution~\cite{YuK15}, residual connection~\cite{Chen2016a} and multi-scale fusion~\cite{Wang2018}.
It demonstrated an advantageous trade-off between receptive field, model complexity and memory consumption. This method also fuses the output of CNNs in three orthogonal views for more robust segmentation of brain tumors.
 In~\cite{Kamnitsas2017a}, an ensemble of multiple models and architectures including DeepMedic~\cite{Kamnitsas2017}, 3D Fully Convolutional Networks (FCN)~\cite{Long2014} and U-Net~\cite{Hefny2015a,Abdulkadir2016} was used for robust brain tumor segmentation. The ensemble method reduces the influence of the meta-parameters of individual CNN models and the risk of overfitting the configuration to a specific training dataset. However, it requires much more computational resources to train and run a set of models.

Training with a large dataset plays an important role for the good performance of deep CNNs. For medical images, collecting a very large training set is usually time-consuming and challenging. Therefore, many works have used data augmentation to partially compensate this problem. Data augmentation applies transformations to the samples in a training set to create new ones, so that a relatively small training set can be enlarged to a larger one. Previous works have used different types of transformations such as flipping, cropping, rotation and scaling training images~\cite{Abdulkadir2016}. In~\cite{Zhang2017}, a simple and data-agnostic data augmentation routine termed \textit{mixup} was proposed for training neural networks. 
Recently, several studies have empirically found that the performance of deep learning-based image recognition methods can be improved by combining predictions of multiple transformed versions of a test image, such as in %data distillation~\cite{Radosavovic2017}, 
pulmonary nodule detection~\cite{Jin2018} and skin lesion classification~\cite{Matsunaga2017}. In~\cite{Isensee2018}, test images were augmented by mirroring for brain tumor segmentation. In~\cite{Wang2018tta}, a mathematical formulation was proposed for test-time augmentation, where a distribution of the prediction was estimated by Monte Carlo simulation with prior distributions of parameters in an image acquisition model. That work also proposed a test-time augmentation-based \textit{aleatoric} uncertainty estimation method that can help to reduce overconfident predictions. The framework in~\cite{Wang2018tta} has been validated with binary segmentation tasks, while its application to multi-class segmentation has yet to be demonstrated.

In this paper, we extend the work of~\cite{Wang17brats} and~\cite{Wang2018tta}, and apply test-time augmentation to automatic multi-class brain tumor segmentation. For a given input image, instead of obtaining a single inference, we augment the input image with different transformation parameters to obtain multiple predictions from the input, with the same network and associated trained weights. The multiple predictions help to obtain more robust inference of a given image. We explore the use of different CNNs as the underpinning network structures. Experiments with BraTS 2018 training and validation set showed that an improvement of segmentation accuracy was achieved by test-time augmentation, and our method can provide uncertainty estimation for the segmentation output.   

\section{Methods}

\subsection{Network Structures}
We explore three network configurations as underpinning CNNs for the brain tumor segmentation task: 1) 3D UNet~\cite{Abdulkadir2016}, 2) the cascaded networks in~\cite{Wang17brats} where a WNet, TNet and ENet was used to segment whole tumor, tumor core and enhancing tumor core respectively, and 3) adapting WNet~\cite{Wang17brats} for one-pass multi-class prediction without using cascaded prediction, which is referred to as multi-class WNet. 

The 3D U-Net has a downsampling and an upsampling path each with four resolution steps. In the downsampling path, each layer has two $3\times 3 \times 3$ convolutions each followed by a Rectified Linear Unit (ReLU) activation function, and then a $2\times 2 \times 2$ max pooling layer was used for downsampling. In the upsamping path, each layer uses a deconvolution with kernel size $2\times 2 \times 2$, followed by two $3\times 3 \times3$ convolutions with ReLU. The network has shortcut connections between corresponding layers with the same resolution in the downsampling path and the upsampling path. In the last layer, a $1\times 1 \times 1$ convolution is used to reduce the number of output channels to the number of segmentation labels, i.e., 4 for the brain tumor segmentation task in the BraTS challenge.

The WNet proposed in~\cite{Wang17brats} is an anisotropic network that considers a trade-off between receptive field, model complexity and memory consumption. It employs dilated convolution~\cite{YuK15}, residual connection~\cite{Chen2016a} and multi-scale prediction~\cite{Wang2018} to improve segmentation performance. The network uses 20 intra-slice convolution layers and four inter-slice convolution layers with two 2D down-sampling layers. Since the anisotropic convolution has a small receptive field in the through-plane direction, multi-view fusion was used to take advantage of the 3D contextual information, where the network was applied in axial, sagittal and coronal views respectively. For the multi-view fusion, the softmax outputs in these three views were averaged. In~\cite{Wang17brats}, WNet is used to segment the whole tumor. TNet for tumor core segmentation uses the same structure as WNet, and ENet for enhancing core segmentation is a variant of WNet that uses only one down-sampling layer. Compared with multi-label prediction, the cascaded networks require longer time for training and testing. To improve the training efficiency, we compare the cascaded networks~\cite{Wang17brats} with the use of multi-class WNet, where a single WNet for multi-label prediction is employed without using TNet and ENet. Therefore, for this variant we change the output channel number from 2 to 4. Multi-view fusion is also used for this multi-class WNet.

\subsection{Data Augmentation for Training and Testing}\label{method_tta}
From the point view of image acquisition, an observed image is only one of many possible observations of the underlying anatomy that can be observed with different spatial transformations and noise. Direct inference with the observed image may lead to a biased result affected by the specific transformation and noise associated with that image. To obtain a more robust prediction, we consider different transformations and noise during the test time. Let $\bm \beta$ and $\bm e$ represent the parameters for spatial transformation and intensity noise respectively. We assume that $\bm \beta$ is a combination of $f_l$, $r$ and $s$, where $f_l$ is a random variable for flipping along each 3D axis, $r$ is the rotation angle along each 3D axis, $s$ is a scaling factor. We consider these parameters following some prior distributions: $f_l \sim Bern(0.5)$, $r\sim U(0, 2\pi)$, $s\sim U(0.8, 1.2)$. For the intensity noise, we assume $ \bm e \sim N(0, 0.05)$ according to the reduced standard deviation of a median-filtered version of a normalized image~\cite{Wang2018tta}.

For data augmentation, we randomly sample $\bm \beta$ and $\bm e$ from the above prior distributions and use them to transform the image. We use the same distributions of augmentation parameters at both training and test time for a given CNN. For test-time augmentation, we obtain $N$ samples from the distributions of $\bm \beta$ and $\bm e$ by Monte Carlo simulation, and the resulting transformed version of the input was fed into the CNN. The $N$ prediction results were combined to obtain the final prediction based on majority voting. 

\subsection{Uncertainty Estimation}
Both model-based (\textit{epistemic}) uncertainty and image-based (\textit{aleatoric}) uncertainty have been investigated for deep CNNs in recent years~\cite{Kendall2017}. The \textit{epistemic} uncertainty is often obtained by Bayesian approximation-based methods such as test-time dropout~\cite{Gal2016}.
In~\cite{Wang2018tta}, test-time augmentation was used to estimate the \textit{aleatoric} uncertainty of segmentation results in a consistent mathematical framework. In this paper, we use test-time augmentation to obtain segmentation results as well as the associated \textit{aleatoric} uncertainty according to~\cite{Wang2018tta}.

The uncertainty estimation is obtained by measuring the diversity of the predictions for a given image. Both the variance and entropy of the distribution can be used to estimate uncertainty. Since variance is not sufficiently representative in the context of multi-modal distributions, we use entropy for the pixel-wise uncertainty estimation desired for segmentation tasks. Let $X$ denote the input image and $Y$ denote the output segmentation. We use $Y^i$ to denote the predicted label for the $i$-th pixel. With the Monte Carlo simulation described in Section~\ref{method_tta}, a set of values for $Y^i$ are obtained $\mathcal{Y}^i = \{y^i_1, y^i_2, ..., y^i_N\}$. The entropy of the distribution of $Y^i$ is therefore approximated as:
\begin{align}\label{eq:entropy_approx_seg}
H(Y^i|X) \approx - \sum_{m=1}^{M} \hat p^i_m \text{ln} (\hat p^i_m)
\end{align} 
where $\hat p^i_m$ is the frequency of the $m$-th unique value in $\mathcal{Y}^i$.

\section{Experiments and Results}

\subsubsection{Data and Implementation Details.}

We used the BraTS 2018\footnote{\url{http://www.med.upenn.edu/sbia/brats2018.html}}~\cite{Menze2015,Bakas2017b,Bakas2017c,Bakas2017,Bakas2018} dataset for experiments. The training set contains images from 285 patients, including 210 cases of HGG and 75 cases of LGG. The BraTS 2018 validation and testing set contain images from 66 and 191  patients with brain tumors of unknown grade, respectively. Each patient was scanned with four sequences: T1, T1ce, T2 and FLAIR. As a pre-processing performed by the organizers, all the images were skull-striped and re-sampled to an isotropic 1mm$^3$ resolution, and the four modalities of the same patient had been co-registered. The ground truth were provided by the BraTS organizers. We uploaded the segmentation results obtained by our method to the BraTS 2018 server, and the server provided quantitative evaluations including Dice score and Hausdorff distance compared with the ground truth.
\begin{figure}[t]
	\centering
	\includegraphics[width=1.0\linewidth]{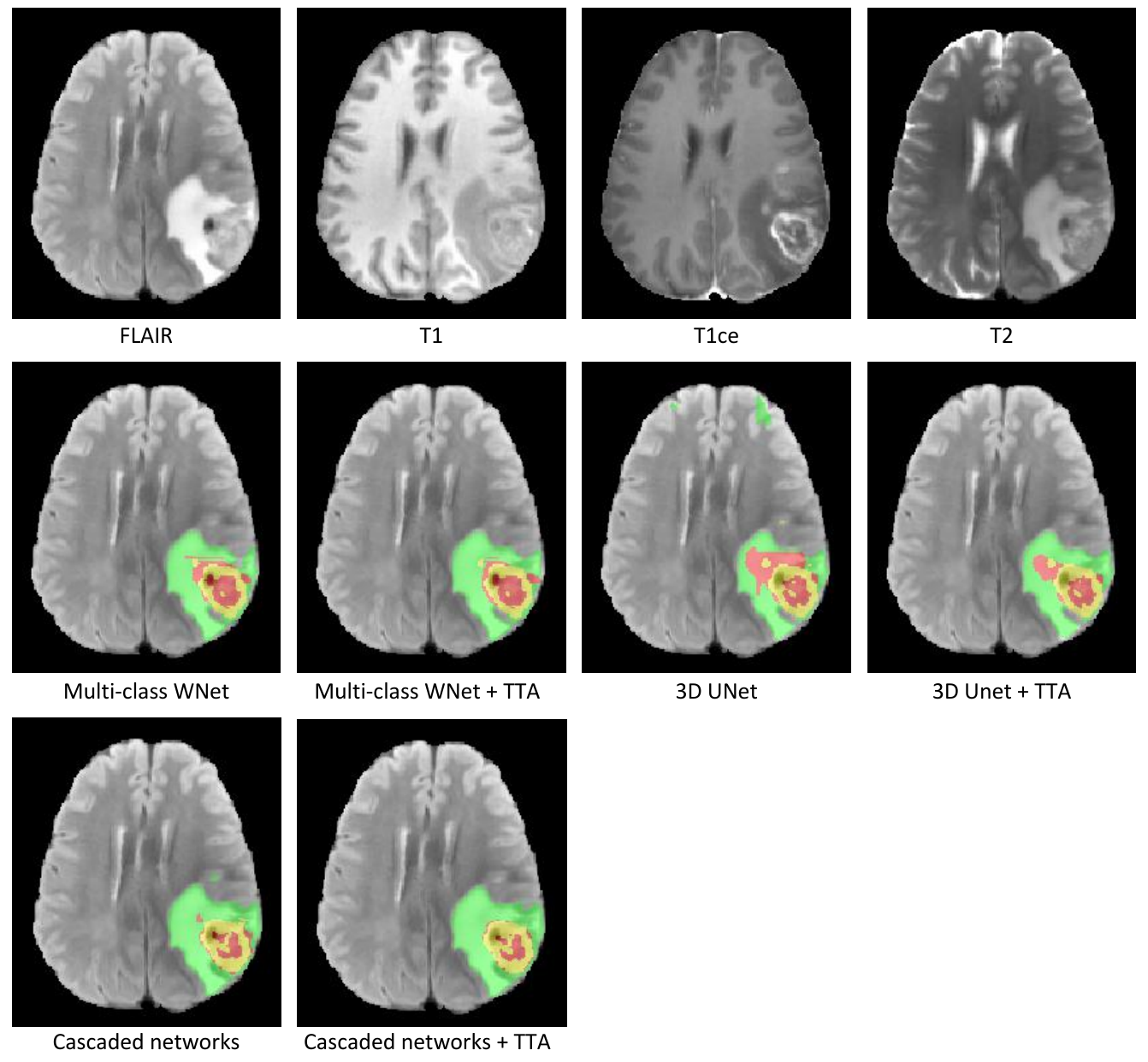}
	\caption[visual comparison]{ 
		An example of brain tumor segmentation results obtained by different networks and test-time augmentation (TTA). The first row shows the four modalities of the same patient. The second and third rows show segmentation results. Green: edema; Red: non-enhancing tumor core; Yellow: enhancing tumor core.
	} 
	\label{fig:visual1}
\end{figure}

We implemented the 3D UNet~\cite{Abdulkadir2016}, multi-class WNet and cascaded networks~\cite{Wang17brats} in Tensorflow\footnote{\url{https://www.tensorflow.org}}~\cite{Abadi2016} using NiftyNet\footnote{\url{http://niftynet.io}}\footnote{\url{https://github.com/taigw/brats18}}~\cite{Gibson2018}. The Adaptive Moment Estimation (Adam)~\cite{Kingma2015} strategy was used for training, with initial learning rate $10^{-3}$, weight decay $10^{-7}$, and maximal iteration 20k. The training patch size was 96$\times$96$\times$96 for 3D UNet and 96$\times$96$\times$19 for multi-class WNet. The batch size was 2 and 4 for these two networks respectively. For the cascaded networks, we followed the configurations in~\cite{Wang17brats}.  The training process was implemented on
an NVIDIA TITAN X GPU. As a pre-processing, each  image was normalized by the mean value and standard deviation. The Dice loss function~\cite{Milletari2016,Fidon2017b} was used for training.  At test time, the augmented prediction number was set to $N = 20$ for all the network structures. The multi-class WNet and cascaded networks were trained in axial, sagittal and coronal views respectively, and the predictions in these three views were fused by averaging at test time. 
 
\subsubsection{Segmentation Results.} 

\begin{figure}[t]
	\centering
	\includegraphics[width=1.0\linewidth]{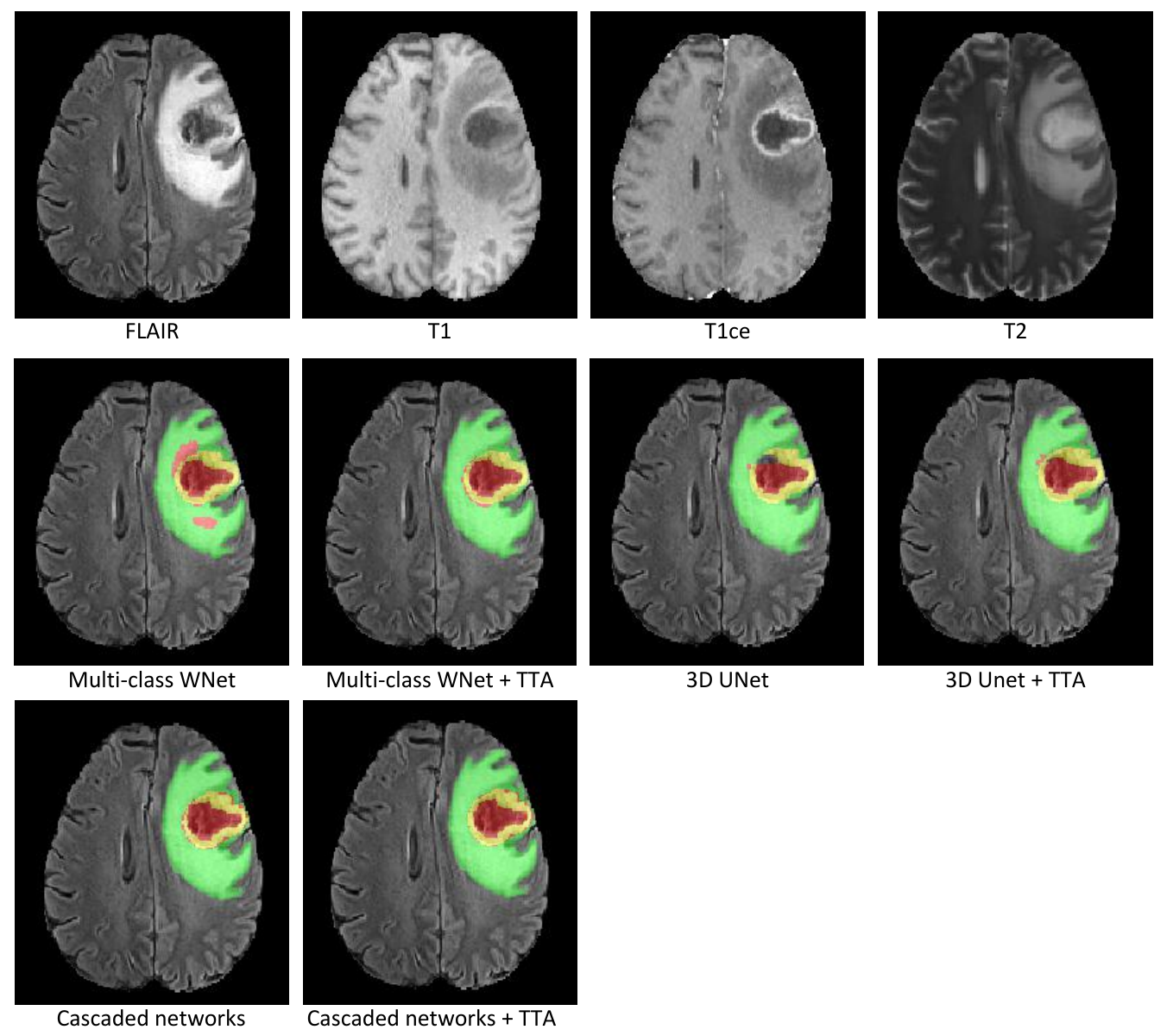}
	\caption[visual comparison]{ 
		Another example of brain tumor segmentation results obtained by different networks and test-time augmentation (TTA). The first row shows the four modalities of the same patient. The second and third rows show segmentation results. Green: edema; Red: non-enhancing tumor core; Yellow: enhancing tumor core.
	} 
	\label{fig:visual2}
\end{figure}

\begin{figure}[t]
	\centering
	\includegraphics[width=1.0\linewidth]{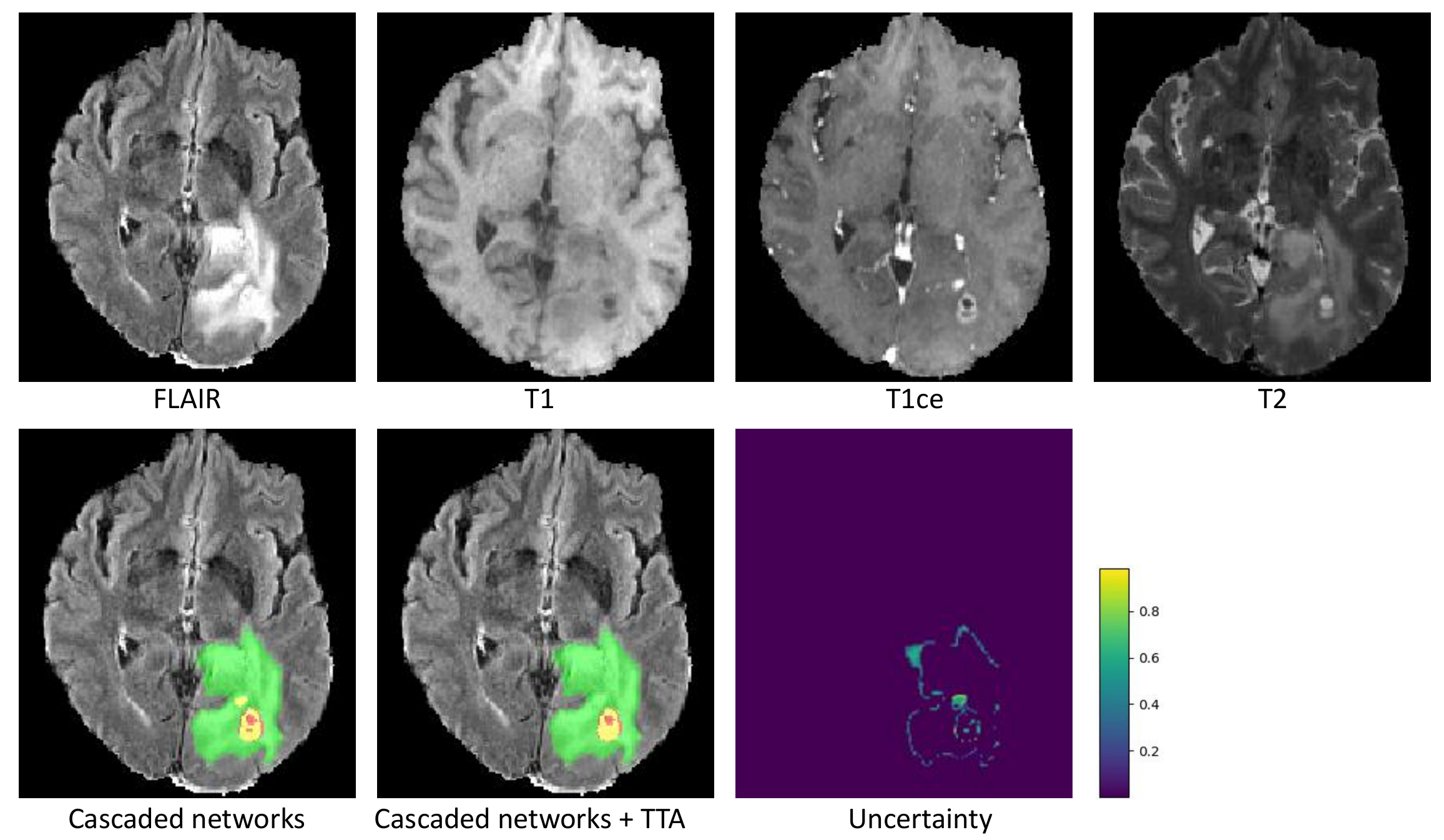}
	\caption[visual comparison]{ 
		An example of segmentation result and uncertainty estimation obtained by cascaded networks~\cite{Wang17brats} with test-time augmentation.
	} 
	\label{fig:visual3}
\end{figure}

Fig.~\ref{fig:visual1} shows an example from the BraTS 2018 validation set. The first row shows the input images of four modalities: FLAIR, T1, T1ce and T2. The second and third rows present the segmentation results of 3D UNet, multi-class WNet, cascaded networks and their corresponding results with test-time augmentation. It can be observed that the initial output of the 3D UNet seems to be noisy with some false positives of edema and non-enhancing tumor core.  After using test-time augmentation, the result becomes more spatially consistent. The output of multi-class WNet also seems to be noisy for the non-enhancing tumor core. A smoother segmentation is obtained by multi-class WNet with test-time augmentation. For the cascaded networks, test-time augmentation also leads to visually better resutls of the tumor core.

Fig.~\ref{fig:visual2} shows another example from the BraTS 2018 validation set.  It can be observed that the 3D UNet obtains a hole in the tumor core, which seems to be an under-segmentation. The hole is filled after using test-time augmentation and the result looks more consistent with the input images. The initial prediction by multi-class WNet seems to have an over segmentation of the non-enhancing tumor core. After using test-time augmentation, the over-segmented regions become smaller, leading to higher accuracy. Test-time augmentation also helps to improve the result of cascaded networks. Fig.~\ref{fig:visual3} shows a case from the BraTS 2018 testing set, where test-time augmentation obtains a better spatial consistency for the tumor core. In addition, it leads to an uncertainty estimation of the segmentation output. It can be observed that most uncertain results focus on the border of the tumor and some potentially mis-segmented regions.

A quantitative evaluation of our different methods on the BraTS 2018 validation set is shown in Table~\ref{tab:valid}.
The initial output of 3D UNet achieved Dice scores of 73.44\%, 86.38\% and 76.58\% for enhancing tumor core, whole tumor and tumor core respectively. 3D UNet with test-time augmentation achieved a better performance than the baseline of 3D UNet, leading to Dice scores of 75.43\%, 87.31\% and 78.32\% respectively.  
For the initial output of multi-class WNet, the Dice score was 75.70\%, 88.98\% and 72.53\% for these three structures respectively. After using test-time augmentation, an improvement was achieved, and the Dice score was 77.70\%, 89.56\% and 73.04\% for these three structures respectively. For the cascaded networks, test-time augmentation leads to higher accuracy for the enhancing tumor core and tumor core. Table~\ref{tab:test} presents the performance of our cascaded networks with test-time augmentation on BraTS 2018 testing set. The average Dice scores for enhancing tumor core, whole tumor and tumor core are 74.66$\%$, 87.78$\%$ and 79.64$\%$, respectively. The corresponding values of Hausdorff distance are 4.16mm, 5.97mm and 6.71mm, respectively. 

\begin{table}
	\centering
	\caption{Mean values of Dice and Hausdorff measurements of different methods on BraTS 2018 validation set. ET, WT, TC denote enhancing tumor core, whole tumor and tumor core, respectively. TTA: test-time augmentation.}
	\label{tab:valid}
	\begin{tabular}{l|c|c|c|c|c|c}
		\hline
		& \multicolumn{3}{c|}{Dice (\%)} & \multicolumn{3}{c}{Hausdorff (mm)}  \\ \hline
		& ET & WT & TC & ET & WT & TC \\ \hline
		3D UNet & 73.44 & 86.38  & 76.58  & 9.37  & 12.00  & 10.37   \\
		3D UNet + TTA & 75.43  & 87.31  & 78.32  & 4.53  & 5.90  & 8.03 \\
		Multi-class WNet & ~75.70~ & ~88.98~  & ~72.53~  & ~4.24~  & ~4.99~  & ~12.13~   \\
		Multi-class WNet + TTA & 77.07  & 89.56  & 73.04  & 4.44  & 4.92  & 11.13 \\
		Cascaded networks & 79.19  & 90.31  & 85.40  & 3.34  & 5.38  & 6.61   \\
		Cascaded networks + TTA & 79.72  & 90.21  & 85.83  & 3.13  & 6.18  & 6.37 \\
		\hline
	\end{tabular}
\end{table}

\begin{table}
	\centering
	\caption{Dice and Hausdorff measurements of our cascaded networks with test-time augmentation on BraTS 2018 testing set. ET, WT, TC denote enhancing tumor core, whole tumor and tumor core, respectively.}
	\label{tab:test}
	\begin{tabular}{l|c|c|c|c|c|c}
		\hline
		& \multicolumn{3}{c|}{Dice (\%)} & \multicolumn{3}{c}{Hausdorff (mm)}  \\ \hline
		& ET & WT & TC & ET & WT & TC \\ \hline
		Mean & ~74.66~ & ~87.78~  & ~79.64~  & ~4.16~  & ~5.97~  & ~6.71~   \\
	Standard deviation & ~25.85~  & ~11.92~  & ~24.97~  & ~7.07~  & ~8.56~  & ~10.27~ \\
		Median &~83.38~ & ~91.33~  & ~89.68~ & ~2.00~  & ~3.32~  & ~3.16~   \\
	25 Quantile & ~72.87~  & ~86.69~  & ~78.24~  & ~1.41~  & ~2.24~  & ~2.00~ \\
		75 Quantile & ~88.64~  & ~94.09~ & ~93.58~  & ~3.00~  & ~5.48~  & ~6.40~ \\
		\hline
	\end{tabular}
\end{table}

\section{Discussion and Conclusion}

For test-time augmentation, we only used flipping, rotation and scaling for spatial transformations. It is also possible to employ more complex transformations such as elastic deformations used in~\cite{Abdulkadir2016}. However, such deformations take longer time for testing and have a lower efficiency. The results show that test-time augmentation leads to an improvement of segmentation accuracy for different CNNs including 3D UNet~\cite{Abdulkadir2016}, multi-class WNet and cascaded networks~\cite{Wang17brats}. Test-time augmentation can be applied to other CNN models as well. The uncertainty estimation obtained by our method can be used for downstream analysis such as uncertainty-aware volume measurement~\cite{Zach2018} and guiding user interactions~\cite{Wang2018}. It would be of interest to assess the impact of test-time augmentation on CNNs trained with state-of-the-art policies such as in~\cite{Isensee2018}. By using test-time augmentation, we investigated the test image-based (\textit{aleatoic}) uncertainty for brain tumor segmentation. It is of interest to investigate how  ensemble of CNNs~\cite{Kamnitsas2017a} can produce \textit{epistemic} uncertainty for this task. For a comprehensive study of uncertainty, it is promising to combine ensemble of models or test-time dropout with test-time augmentation. This will be left for future work.

In conclusion, we explored the effect of test-time augmentation on CNN-based brain tumor segmentation. We used 3D U-Net, 2.5D multi-class WNet and cascaded networks as the underpinning network structures. For training and testing, we augmented the image by 3D rotation, flipping, scaling and adding random noise. Experiments with BraTS 2018 training and validation set
show that test-time augmentation helps to improve the brain tumor segmentation accuracy for different CNN structures and obtain uncertainty estimation of the segmentation results.

\subsubsection{Acknowledgements.}\label{sec:acknowledgements}
We would like to thank the NiftyNet team. This work was supported through an Innovative Engineering for Health award by the Wellcome Trust [WT101957,  WT97914, 203145Z/16/Z, 203148/Z/16/Z], Engineering and Physical Sciences Research Council (EPSRC) [NS/A000027/1, NS/A000049/1, NS/A000050/1], the National Institute for Health Research University College London Hospitals Biomedical Research Centre (NIHR BRC UCLH/UCL High Impact Initiative), hardware donated by NVIDIA, and the Health Innovation Challenge Fund [HICF-T4-275].
%\begin{thebibliography}{5}
\bibliographystyle{splncs03}
\bibliography{./reference/miccai2017,./reference/miccai2018,./reference/midl2018}
%\bibliography{reference/miccai2016}

%\end{thebibliography}

%
% second contribution with nearly identical text,
% slightly changed contribution head (all entries
% appear as defaults), and modified bibliography
%

\end{document}